\documentclass[runningheads]{llncs}
\usepackage{graphicx}
\begin{document}
\title{Federated Learning for Drowsiness Detection in Connected Vehicles}
%
%
\author{William Lindskog\inst{1, 2}
\and
Valentin Spannagl\inst{2}
\and
Christian Prehofer\inst{1,2}
}
\authorrunning{W. Lindskog et al.}
%
\institute{DENSO Automotive Deutschland GmbH, Freisinger Str. 21, 85386, Eching \email{w.lindskog@eu.denso.com}\\ \and
Technical University of Munich, Germany
}
\maketitle              
%

\begin{abstract}
Ensuring driver readiness poses challenges, yet driver monitoring systems can assist in determining the driver's state. By observing visual cues, such systems recognize various behaviors and associate them with specific conditions. For instance, yawning or eye blinking can indicate driver drowsiness. Consequently, an abundance of distributed data is generated for driver monitoring. Employing machine learning techniques, such as driver drowsiness detection, presents a potential solution. However, transmitting the data to a central machine for model training is impractical due to the large data size and privacy concerns. Conversely, training on a single vehicle would limit the available data and likely result in inferior performance. To address these issues, we propose a federated learning framework for drowsiness detection within a vehicular network, leveraging the YawDD dataset. Our approach achieves an accuracy of 99.2\%, demonstrating its promise and comparability to conventional deep learning techniques. Lastly, we show how our model scales using various number of federated clients.
\end{abstract}

\keywords{Federated Learning  \and Driver Drowsiness \and Connected Vehicles}
%
%
%




\section{Introduction}\label{sec:introduction}

Road accidents are predominantly caused by human errors, accounting for 90\% of incidents in the United States in 2015 \cite{dingus2016driver}. While drivers and passengers benefit from the safety features of vehicles, vulnerable road users such as cyclists and pedestrians remain at greater risk. Drowsy driving is a prevalent issue, with a significant number of accidents attributed to this cause. 
The vision of fully automated vehicles offers potential solutions to address the problem of driver drowsiness. However, achieving fully automated driving remains a future aspiration, and the current trajectory suggests the adoption of shared driver-machine models \cite{fridman2018human}. For complex driving scenarios, the driver is required to assume control of the vehicle, necessitating a state of readiness.

Driver readiness is contingent upon various factors and influenced by the driver's state \cite{mioch2017driverreadiness}. Automotive companies have developed different types of Driver Monitoring Systems (DMS) to enhance road safety \cite{halin2021survey}\cite{sikander2018driver}. These systems analyze driver behavior and appearance to detect signs of hazardous conditions, such as distraction or drowsiness, indicating when the driver may not be prepared to assume control \cite{cancello2022detection}. However, interpreting these signs can be challenging and subject to multiple interpretations. Consequently, Driver Monitoring encompasses several domains, with Driver Drowsiness Detection being one of them.

Various approaches have been employed to detect driver fatigue. Direct physiological measurements, involving the use of body sensors to track metrics like heart rate, have yielded promising results. Another approach utilizes vehicle sensors to monitor parameters like steering wheel angle, detecting anomalous patterns \cite{sahayadhas2012detecting}. A third approach combines direct assessment of driver behavior with non-intrusive sensors. Optical algorithms, coupled with Deep Learning techniques, extract the driver's state from camera recordings, demonstrating improved success rates \cite{fridman2016gaze}. However, in-vehicle Driver Monitoring presents challenges due to the large video data size and privacy constraints. 

Federated Learning (FL), a Deep Learning methodology, offers a privacy-aware solution to train Machine Learning models on distributed data. By only transmitting the model parameters through the federated network, rather than raw data, FL reduces message sizes and minimizes the potential attack surface for adversarial attacks. FL has started to see real-world implementations, mainly in medicine \cite{rieke2020future}, but is being applied in other industries such as automotive \cite{lindskog2022federated}. 

In order to tackle these challenges, our study introduces an FL framework tailored for drowsiness detection in a vehicular network, employing the YawDD dataset. Remarkably, our approach attains a remarkable accuracy rate of 99.2\%, showcasing its potential and comparability to conventional deep learning methods. 
Our main contributions are: 
\begin{itemize}
    \item Federated Learning framework for driver drowsiness detection using YawDD dataset for processing single frames and sequences. 
    \item With our evaluation, we show how model performance scales when increasing the number of federated clients. 
    \item We achieve great results of 99.2\% when classifying normal driving, talking and yawning driver. 
\end{itemize}
\section{Related Work}\label{sec:related_work}

Driver Drowsiness Detection is an actively researched area, and numerous studies have been conducted to address this critical issue. \cite{abtahi2014yawdd} developed a benchmark implementation using the YawDD dataset, focusing on in-vehicle applications. However, the accuracy of their model is limited.
It is evident that there is room for improvement in achieving higher detection accuracies for driver drowsiness.


In recent works, \cite{salman2021driver} proposed a Convolutional Neural Network (CNN) architecture on the YawDD dataset, utilizing ensemble learning to achieve an impressive accuracy of approximately 99\%. \cite{rajkar2022driver} employed a simple CNN with dropout technique, obtaining an average accuracy of 96\% on YawDD by focusing on the eyes and mouth and using labels of open or closed states for these facial features. \cite{junaedi2018driver} introduced the PERCLOSE formulation, utilizing the eye area to determine drowsiness on the YawDD dataset. However, this approach encounters challenges when the driver's face or eyes are not detectable. \cite{al2020yawn} explored the application of Recurrent Neural Networks (RNNs) on the YawDD dataset, achieving an accuracy of 96\%.

The aforementioned studies predominantly employ conventional deep learning techniques, such as ensemble learning, CNNs, and RNNs, to attain high accuracies in driver drowsiness detection. However, the integration of federated learning in the driver monitoring field remains underrepresented. \cite{zafar2021federated} presented one of the few works focusing on drowsiness detection using FL. They proposed a two-stage approach, utilizing the PERCLOSE method and another drowsiness metric called FOM, to identify fatigue. The application of the federated learning strategy called Dynamic Averaging yielded promising results, and the performance evaluation was conducted on the NTHU dataset. \cite{zhang2022privacy} also used the NTHU dataset and the YawDD dataset. They present a privacy-preserving federated transfer learning method called PFTL-DDD for detecting driver drowsiness. The proposed method uses fine-tuning transfer learning on the FL system's initial model and a CKKS-based security protocol to encrypt exchanged parameters, protecting driver privacy. The results show that the method is more accurate and efficient compared to conventional FL methods and reduces communication costs. Nevertheless, they do not include how many clients the evaluate their models with. \cite{li2021federated} also proposed a federated transfer learning model, but applied it to construction workers and fatigue monitoring. Lastly, \cite{yang2022efficient} proposed an asynchronous federated scheme in internet of vehicles. They evaluate their model EHAFL on YawDD dataset and show how their model can reduce communication costs with 98\% with a slight decrease in accuracy. 

While conventional deep learning techniques have demonstrated success in driver drowsiness detection, the potential of FL in this domain remains largely untapped. Moreover, the studies that have investigated FL for DMS do not show how performance scales when number of participating clients is increased.

\section{YawDD Dataset}\label{subsec:(METHOD)_YawDD_dataset}
\begin{figure}[h!]
    \centering
    \includegraphics[width=\textwidth]{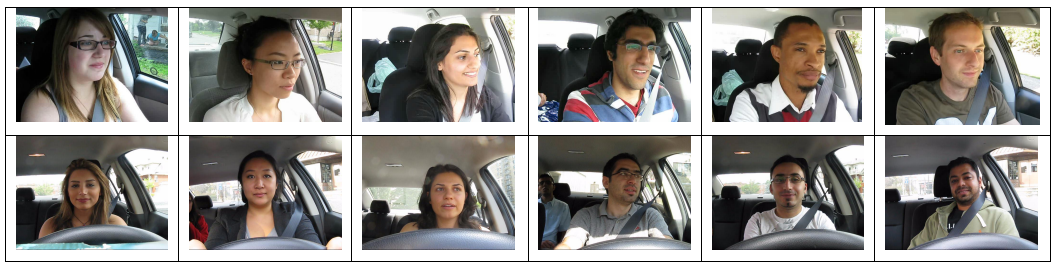}
    \caption{Samples from the YawDD dataset in two perspectives. Top: Rear mirror. Bottom: Dash \cite{abtahi2014yawdd}}
    \label{fig:yawddSamples}
\end{figure}

The YawDD dataset, as documented by \cite{abtahi2014yawdd}, comprises video recordings of drivers exhibiting various behaviors, with particular emphasis on yawning as a key indicator of drowsiness. To establish a baseline, normal driving videos were included to depict drivers in an alert state, free from drowsiness. Another category of videos captured drivers engaged in conversation. In a binary classification task involving yawning and normal driving, the act of mouth opening serves as an indication of yawning behavior. However, the introduction of talking data poses a challenge as this behavior can no longer be solely relied upon. The YawDD dataset consists of two subsets: one containing 322 videos and the other with 29 videos. The larger subset captures drivers from a rear mirror perspective, while the smaller subset features a camera placed in front of the driver on the dashboard, see Figure \ref{fig:yawddSamples}. A total of 107 drivers (57 male and 50 female) were recorded, with each driver providing at least three videos for each behavior category. Additionally, the YawDD dataset includes recordings with occlusions such as sunglasses or scarves. 

\subsection{Data Preprocessing}\label{subsubsec:(METHOD)_data_preprocessing}
Given that the provided labels in the YawDD dataset were different, encompassing a few seconds before and after the actual yawning event in a yawning video, we performed frame-level labeling of the entire dataset to ensure precise classification for this study. Analyzing the dataset is crucial to identify its properties. The dataset is divided into two subsets based on camera perspectives: rear mirror (320 items) and dash (29 items). Despite both perspectives being realistic, the larger rear mirror set is selected for further analysis. The file names reveal a potential issue: some videos have dual labels. For instance, a sample video includes footage of talking and yawning. Upon reviewing sample videos, labeling appears more problematic as a video labeled "Yawning" contains a yawning event within normal driving. The drivers are throughout the videos driving normally, talking or yawning; thus we tackle a 3-class classification task. 

\begin{figure}[h!]
    \centering
    \includegraphics[width=\textwidth]{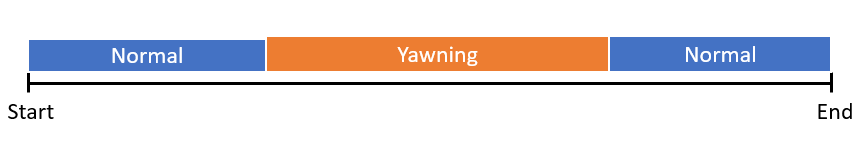}
    \caption{Sample includes two categories. }
    \label{fig:sample_categories}
\end{figure}

Resizing is the subsequent step for data reduction while preserving crucial information. The camera used in \cite{abtahi2014yawdd} captures YawDD data at a resolution of 640x480 pixels, resulting in 307,200 pixel values per color layer. In the RGB format, this equates to 921,600 values per frame. Deep learning memory consumption scales with input data size, so minimizing input sample size is preferable. Lowering resolution causes image structures to blur. At 80x80 pixels, objects become indistinct, such as glasses merging with the eye structure. Resolutions above 160x160 pixels are recommended, with 320x320 already offering improvement compared to the original. The last step in frame pre-processing involves color representation. Since the driver drowsiness detection approach relies on behavioral measures, driver actions play a vital role. Yawning and eye blinking, identified as indicators, are not influenced by color representation. Grayscaling converts RGB to grayscale, significantly reducing data size.

\section{Method for Driver Drowsiness}\label{sec:method}
In this section, deep learning for spatial-temporal data and the tools applied in our study. 

\subsection{Deep Learning and Spatial-Temporal Data}\label{subsec:(METHOD)_deep_learning_and_spatial_temporal_Data}
Sequential data as input for the neural network includes the concept of spatial-temporal data. Images can be considered spatial data, where pixels correspond to locations, areas, and distances. When single frames act as input, the neural network relies on spatial information, and the task is typically image classification using CNNs. However, in YawDD, the original data is in video form, introducing temporal information. The order of frames in a video is not arbitrary, and the relation between consecutive frames is temporal. With sequences of frames, we have spatial-temporal information, and the task becomes sequence or video classification. Utilizing spatial-temporal information is desirable as it preserves an additional dimension of information. For sequential data, RNNs are a specialized variation of deep learning. \cite{ed2020real} used an RNN extension of CNNs known as 3D CNNs, which extend CNNs to the third dimension, representing time. They achieved promising results, outperforming other methods like LSTMs. There is still nevertheless a discussion whether to use 2D or 3D CNNs for video processing \cite{chen2021deep}. Using single frames instead of sequences can help overcome hardware constraints and reduce the number of parameters to consider. Processing single frames could however result in a loss of information. Thus, we evaluate both 2D and 3D CNN on YawDD dataset.


\subsection{Federated Learning Tools}
\begin{figure}[h!]
    \centering
    \includegraphics[width=0.8\textwidth]{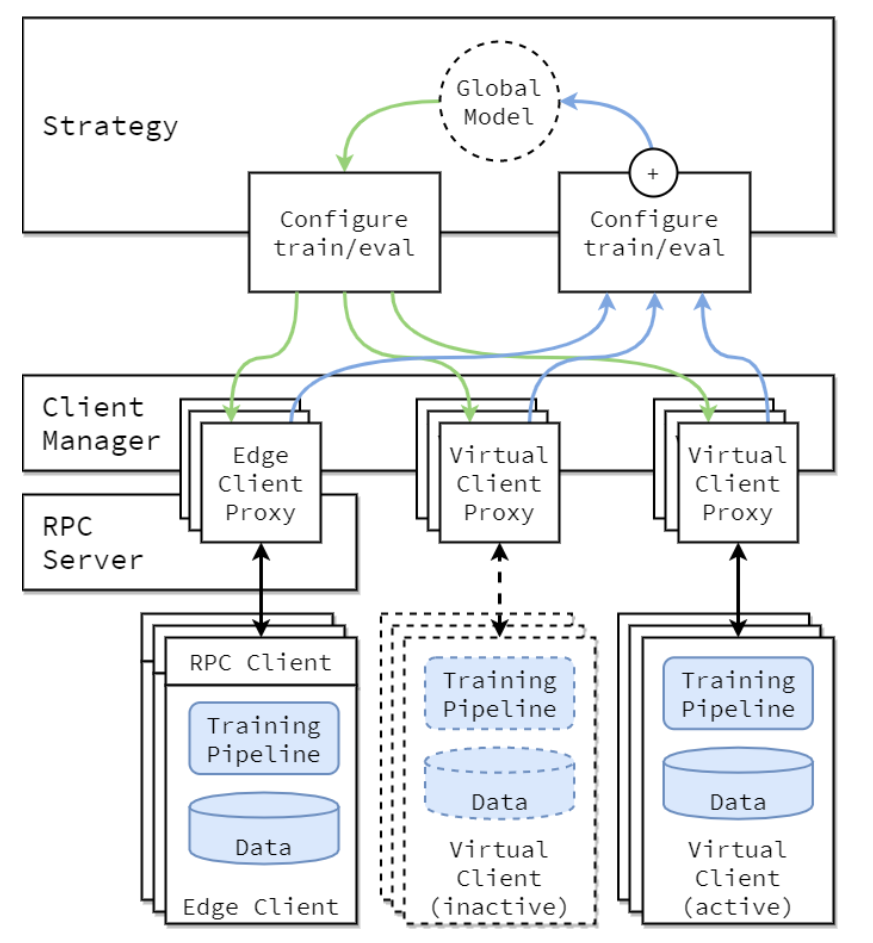}
    \caption{Flower core framework architecture \cite{beutel2020flower}.}
    \label{fig:Flower_architecture}
\end{figure}

Our FL architecture was implemented using the Flower framework \cite{beutel2020flower} in conjunction with PyTorch. The Flower framework provides three key scripts: the \emph{server} or \emph{main} script, responsible for managing the federated training process; the \emph{client} script, which handles local training and client-side evaluation; and the \emph{utils} script, which incorporates essential functionalities such as data loading. Figure \ref{fig:Flower_architecture} depicts the Flower core framework architecture, how server and clients communicate and the possibility to simulate clients using a built-in virtual engine. 

PyTorch offers a straightforward approach to load an image dataset from a folder. The \emph{ImageFolder} function facilitates the loading and transformation of image samples along with their corresponding labels. Transformations are composed within the \emph{transform} object, which includes operations such as grayscale conversion, conversion to Tensor format, normalization, and standardization. The label information is inferred from the folder structure, as the dataset folder comprises three subfolders for each class.

After loading the entire dataset into a variable, the next step involves splitting and assigning the subsets. We utilize the \emph{random\_split} function, which randomly permutes the dataset and splits it according to the specified length array. To ensure reproducibility, the random permutation is generated using a seeded number. Importantly, the randomization process respects the proportions of the label collections, guaranteeing that both splits maintain the same class distribution. 

\section{Evaluation and Results}\label{sec:results}
The experiments were conducted on the Ubuntu distribution, which is built on the Linux operating system. Specifically, Ubuntu version 22.04 was utilized for this study. The hardware setup consisted of an AMD seventh-generation processor and a NVIDIA GeForce RTX 3070 graphics card.

At first, the loaded dataset size is controlled and the size 127.887 matches the number of files in the dataset folder. The 90:10 split is validated with a training set, containing 90\% of the data, and a test set, which possesses 10\% of the data. We also specify relevant hyperparameters before training. We search for optimal values for these and illustrate our results using the best choices found. The hyperparameter search space can be found in Table \ref{tab:(RESULT)_hyperparameters}. Moreover, we use FedAvg \cite{konevcny2016federated} as a federated strategy when aggregating the client updates at the server. Adam \cite{kingma2014adam} is set as an optimizer. 

\begin{table}[h!]
\caption{Hyperparameter search space. \texttt{sequence\_length} and \texttt{frame\_skipping} parameters are only relevant for 3D-CNNs. }\label{tab:(RESULT)_hyperparameters}
\centering
\begin{tabular}{|c|c|}
\hline
\textbf{Hyperparameter} &  Search Space Values\\
\hline
\texttt{learning\_rate} & $\{0.0001, 0.001, 0.002, 0.005, 0.01, 0.1\}$\\
\texttt{momentum} & $\{0.01, 0.02, 0.05, 0.1, 0.2, 0.5, 0.9\}$ \\
\texttt{batch\_size} & $\{2, 8, 16, 32, 64, 128 \}$ \\
\texttt{weight\_decay} & $\{0.0001, 0.001, 0.002, 0.005, 0.01, 0.1\}$ \\
\texttt{nbr\_clients} & $\{2, 4, 8, 16, 20, 40 \}$ \\
\texttt{sequence\_length} & $\{8, 10, 12, 14, 26, 18, 20 \}$ \\
\texttt{frame\_skipping} & $\{2, 3, 4, 5, 6, 8, 10, 12 \}$\\
\hline
\end{tabular}
\end{table}

\subsection{3D-CNN and Video Sequence Processing}
\begin{figure}[h!]
    \centering
    \includegraphics[width=\linewidth]{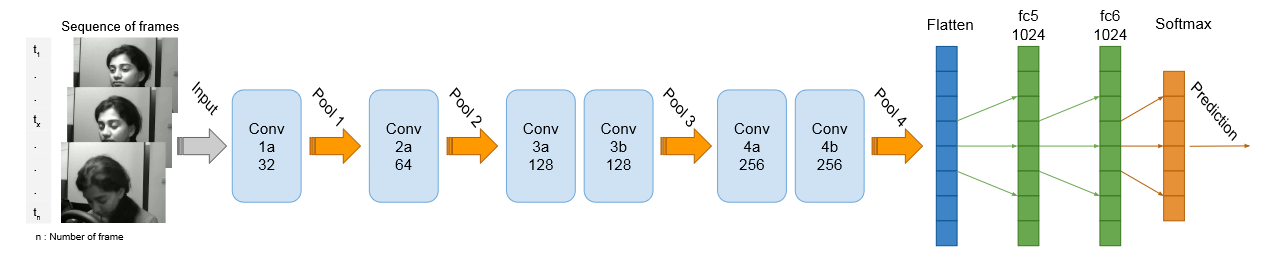}
    \caption{3D-CNN architecture as in \cite{ed2020real}}
    \label{fig:3d_architecture}
\end{figure}

Based on \cite{ed2020real} work, the hyperparameter values were adopted and a model was constructed as in Figure \ref{fig:3d_architecture}. 
The sequence length was found to influence the results and an initial sequence length of 16 was used, but other initializations did not improve performance. Frame skipping, introduced to enhance temporal information, showed no effect, with a commonly used frame skipping value of five. The batch size, limited by memory, was set to 2 due to the sequence length. The learning rate was the only parameter that showed some change, with smaller rates resulting in slower convergence but reduced fluctuations. The best values for frame skipping and sequence length were used after hyperparameter tuning. Highest accuracy achieved was 90.1\% and processing was slow. To improve the approach and achieve better results, the process can be streamlined and simplified. Using single frames instead of sequences can help overcome hardware constraints and reduce the number of parameters to consider. 



\subsection{2D-CNN and Image Processing}
\begin{figure}[h!]
    \centering
    \includegraphics[width=\linewidth]{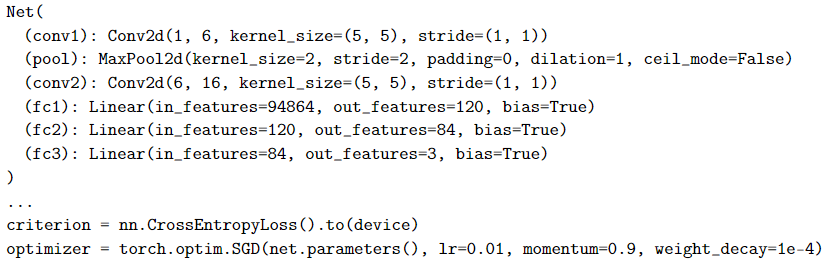}
    \caption{PyTorch code for convolutional neural network we use. The parameter values for the optimizer are initial values. }
    \label{fig:neural_network_code}
\end{figure}

The adopted model is a 2D-Convolutional Neural Network (CNN) for image classification, in which we process the videos frame-by-frame. We draw inspiration from the work of \cite{ed2020real} when constructing the 2D-CNN and the architecture can be found in Figure \ref{fig:neural_network_code}

\begin{figure}[h!]
    \centering
    \includegraphics[width=0.86\textwidth]{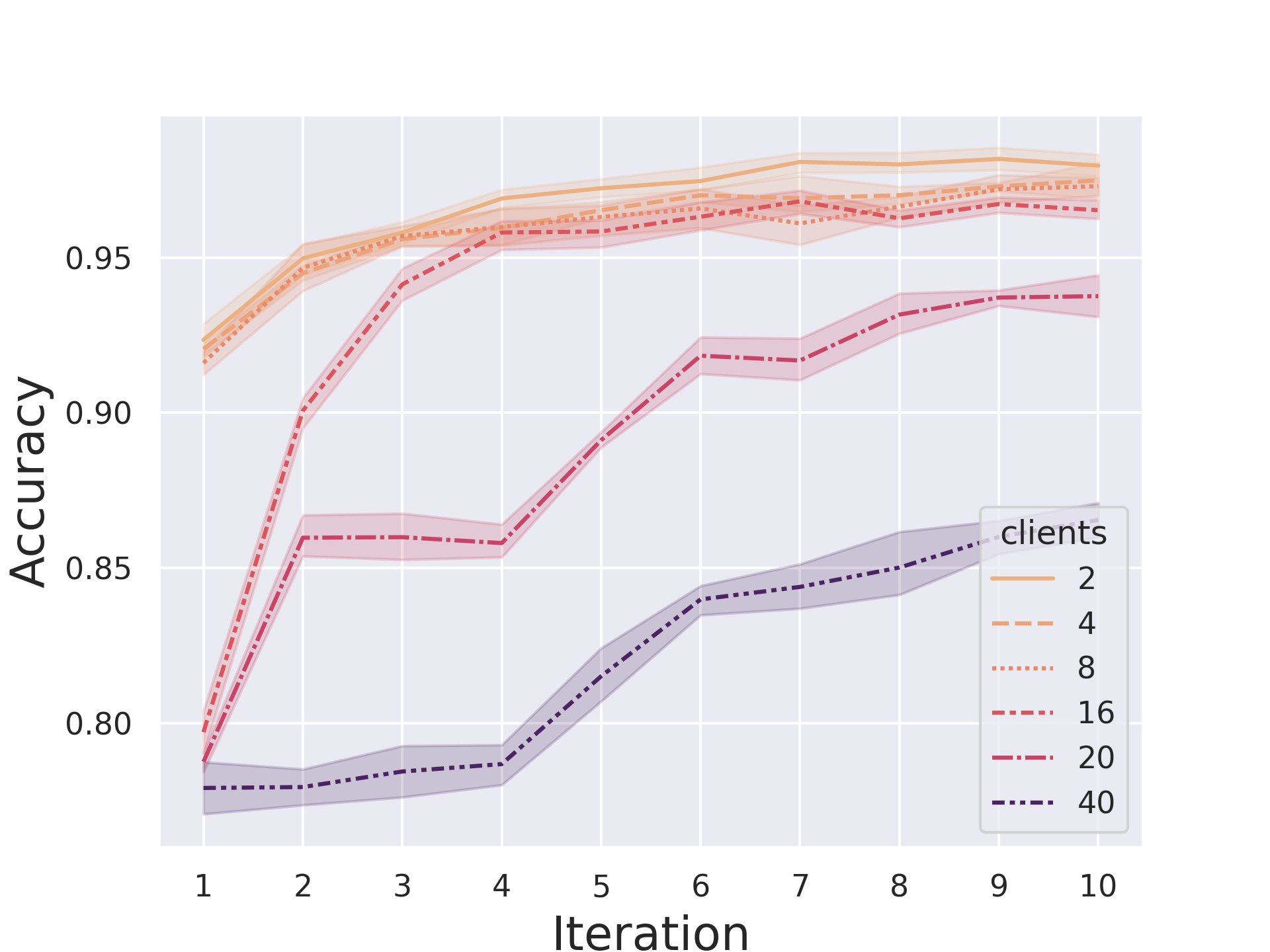}
    \caption{Test accuracy for various number of clients. Accuracy is shown with significant intervals, averaged over 5 runs. }
    \label{fig:(RESULTS)_accuracies}
\end{figure}

\begin{figure}[h!]
    \centering
    \includegraphics[width=0.85\textwidth]{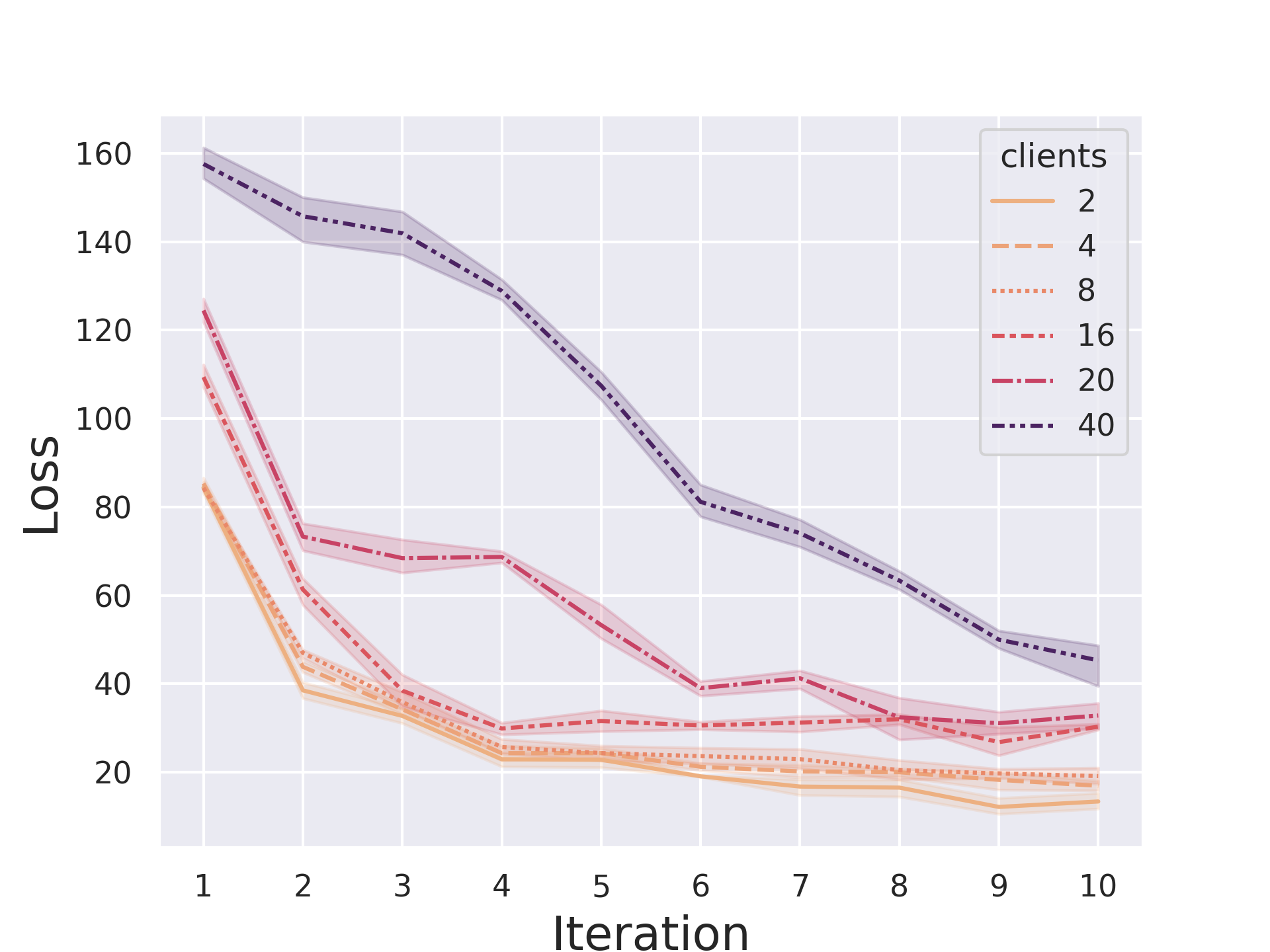}
    \caption{Test categorical cross-entropy loss for various number of clients. Accuracy is shown with significant intervals, averaged over 5 runs. }
    \label{fig:(RESULTS)_loss}
\end{figure}

We thereafter evaluate our model using the mentioned pre-processing steps on a frame-by-frame basis. The task is a 3-class classification task in which we seek to correctly classify drivers driving normally with closed mouth, drivers talking, and drivers yawning. We illustrate our results using test accuracy and categorical cross-entropy loss. The values illustrated in Figure \ref{fig:(RESULTS)_accuracies} and \ref{fig:(RESULTS)_loss} are averaged scores over 5 runs. We also show respective significance interval for both illustrations.

\begin{figure}[h!]
    \centering
    \includegraphics[width=\textwidth]{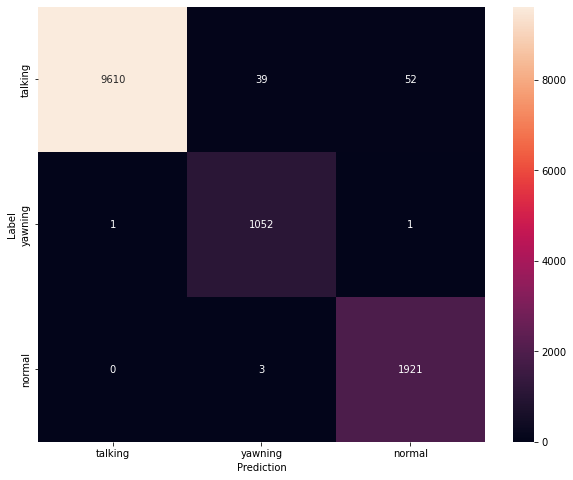}
    \caption{Confusion matrix with predictions for driver drowsiness. }
    \label{fig:(RESULTS)confusion_matrix}
\end{figure}

The test accuracy in Figure \ref{fig:(RESULTS)_accuracies} shows that highest accuracy is achieved using 2 clients. Maximum accuracy achieved is 99.2\% and prediction outcomes can be seen in Figure \ref{fig:(RESULTS)confusion_matrix}. We notice a slight decrease in performance when increasing the number of participating federated clients. This is clearly shown when increasing the number of clients from $16 \rightarrow 20$ and $20 \rightarrow 40$. However, the performance is fairly stable between $2 \rightarrow 16$ clients. The optimal choice of hyperparameters seems to be a low learning rate, weight decay, number of participating federated clients and batch size. A larger momentum gives better results. 

In Figure \ref{fig:(RESULTS)confusion_matrix}, we show the predictions of our best run for driver drowsiness using 2 clients. Similar pattern is seen when increasing the number of participating clients. From the results, we read that our model can easily distinguish the classes yawning and and normal driving and that talking is sometimes mistaken for yawning or normal driving, almost in same proportion.  

\section{Discussion}\label{sec:discussion}
Firstly, we controlled the loaded dataset size, which contained 127,887 files, ensuring that the dataset folder was accurately represented. We split the dataset into a training set, comprising 90\% of the data, and a test set, containing the remaining 10\%. This splitting strategy is commonly employed in machine learning to evaluate the generalization performance of models. By using this approach, we can assess how well our model performs on unseen data. To ensure optimal performance, we conducted a hyperparameter search to identify the best values for the relevant parameters. 


In our experiments, we employed the FedAvg FL strategy, as proposed by \cite{konevcny2016federated}. This strategy enables efficient aggregation of client updates at the server while preserving data privacy. In other studies, researchers can come to evaluate other federated strategies such as FedProx \cite{li2020federated} which could serve the area of personalized FL \cite{tan2022towards} better in case one seeks to split unique user data to individual clients. 
We first evaluate our 3D-CNN on YawDD data and achieve 90.1\% accuracy. This is not as good as results in other studies. As mentioned, using single frames for processing can help overcome hardware limitations and reduce the number of parameters considered. 

For evaluation purposes, we applied the specified pre-processing steps on a frame-by-frame basis. Our task involved a 3-class classification, where the objective was to accurately classify drivers engaged in normal driving with a closed mouth, drivers who were talking, and drivers who were yawning. To assess the performance of our model, we measured the test accuracy and the categorical cross-entropy loss. The results, shown in Figure \ref{fig:(RESULTS)_accuracies} and Figure \ref{fig:(RESULTS)_loss}, respectively, were averaged over 5 runs to account for potential variations. Analyzing the test accuracy results presented in Figure \ref{fig:(RESULTS)_accuracies}, we observe that the highest accuracy of 99.2\% was achieved when using 2 clients. This finding indicates that a smaller number of participating federated clients yielded superior performance. However, we notice a slight decrease in accuracy when the number of clients increased from 16 to 20 and from 20 to 40. This decline in performance suggests that as more clients participate, the aggregation process becomes more challenging, potentially due to increased heterogeneity or more likely, size of local datasets becoming too small. With smaller local datasets, the likelihood of it including sufficient and representative data decreases and thus we may experience an increase in heterogeneity. Heterogeneous datasets or non-identical and independent (non-IID) datasets are prevalent in FL and researchers have studied this extensively \cite{zhao2018federated}. 

Interestingly, the performance remained relatively stable when the number of clients ranged from 2 to 16. This observation implies that a moderate number of participating clients is optimal for the task at hand. To further improve the model's performance, we identified several key hyperparameters that played a crucial role. These include a low learning rate, weight decay, number of participating federated clients, and batch size. Additionally, we found that a larger momentum value yielded better results, indicating the importance of effectively leveraging momentum during the optimization process.

To gain more insights into the classification results, we analyzed the confusion matrix shown in Figure \ref{fig:(RESULTS)confusion_matrix}. This matrix represents the predictions obtained from our best run using 2 clients. Notably, similar patterns were observed when increasing the number of participating clients. From the confusion matrix, we deduce that our model can effectively distinguish between yawning and normal driving classes. However, there is a notable confusion between the talking class and the yawning class, as well as between the talking class and the normal driving class. These misclassifications suggest that drivers who are talking exhibit certain facial movements or patterns that resemble both yawning and normal driving. Since we are operating on a frame-per-frame level, there will be certain cases where the decision boundary is "blurry", i.e., cases which look alike but belong to different classes. Further investigation into the distinguishing features between these classes could potentially lead to improvements in the model's performance. To extend this study, future research could focus on exploring additional feature engineering techniques or investigating more advanced models to further enhance the classification accuracy. Additionally, collecting more diverse and extensive datasets could provide a more comprehensive evaluation of the model's performance in real-world scenarios.

One interesting area of research is the field of Personalized FL \cite{tan2022towards}. We see that researchers can apply learnings from this field onto the problem of accurate DMS. This includes investigating different model architectures and aggregating algorithms e.g. FL with personalization layers \cite{arivazhagan2019federated}. 
\section{Conclusion}\label{sec:conclusion}
While most of the vehicle control is handled by machines, drivers still need to be prepared to handle complex situations. Overcoming the challenges of ensuring driver readiness is crucial, and driver monitoring systems play a significant role in assessing the driver's state. These systems utilize visual cues to recognize various behaviors and associate them with specific conditions, such as drowsiness indicated by yawning or eye blinking. Consequently, an abundance of distributed data is generated for driver monitoring.

To address the task of driver drowsiness detection, machine learning techniques, such as the one employed in this study, offer a potential solution. However, transmitting the vast amount of data to a central machine for model training is impractical due to privacy concerns and the sheer size of the data. On the other hand, training the model solely on a single vehicle would limit the available data and likely result in inferior performance.

To overcome these challenges, we propose an FL framework within a vehicular network for drowsiness detection, utilizing the YawDD dataset. Our approach demonstrates impressive accuracy, achieving a rate of 99.2\%. This result highlights the promise and comparability of our method to conventional deep learning techniques. Our main contributions are: 
\begin{itemize}
    \item Federated Learning framework for driver drowsiness detection using YawDD dataset for processing single frames and sequences.  
    \item With our evaluation, we show how model performance scales when increasing the number of federated clients. 
    \item We achieve great results of 99.2\% when classifying normal driving, talking and yawning driver. 
\end{itemize}

\bibliographystyle{splncs04}
\bibliography{main.bib}
%




\end{document}